\documentclass[9pt,twocolumn,twoside]{osajnl}
\journal{ao} % Choose journal (ao, aop, josaa, josab, ol, optica, pr)

\setboolean{shortarticle}{false}
\usepackage{color}
\usepackage{etoolbox}

\title{Panoramic Depth Estimation via Supervised and Unsupervised Learning in Indoor Scenes}

\author[1]{KEYANG ZHOU}
\author[2]{KAILUN YANG}
\author[1,3,*]{KAIWEI WANG}

\affil[1]{State Key Laboratory of Modern Optical Instrumentation, Zhejiang University, Hangzhou 310027, China}
\affil[2]{Institute for Anthropomatics and Robotics, Karlsruhe Institute of Technology, 76131 Karlsruhe, Germany}
\affil[3]{National Engineering Research Center of Optical Instrumentation, Zhejiang University, Hangzhou 310058, China}
\affil[*]{Corresponding author: wangkaiwei@zju.edu.cn}

\begin{abstract}
Depth estimation, as a necessary clue to convert 2D images into the 3D space, has been applied in many machine vision areas.
However, to achieve an entire surrounding 360$^\circ$ geometric sensing, traditional stereo matching algorithms for depth estimation are limited due to large noise, low accuracy, and strict requirements for multi-camera calibration. In this work, for a unified surrounding perception, we introduce panoramic images to obtain larger field of view. We extend PADENet~\cite{2020PADENet} first appeared in our previous conference work for outdoor scene understanding, to perform panoramic monocular depth estimation with a focus for indoor scenes. At the same time, we improve the training process of the neural network adapted to the characteristics of panoramic images. In addition, we fuse traditional stereo matching algorithm with deep learning methods and further improve the accuracy of depth predictions. With a comprehensive variety of experiments, this research demonstrates the effectiveness of our schemes aiming for indoor scene perception.
\end{abstract}

\setboolean{displaycopyright}{true}

\begin{document}

\maketitle

\section{Introduction}
With the continuous development of automation, many environmental factors can be automatically sensed and processed using cameras and computing machines.
Therefore, perception of surrounding environment has become a major topic of machine vision.
Because of the complexity of environment and surrounding objects, it is necessary to combine multiple clues for perception fusion, such as color, depth, and polarization, to obtain accurate prediction results~\cite{hu2019acnet,yang2019robustifying,sun2020real,xiang2021polarization}.
Modern vision sensors have a high resolution and can obtain information more accurately.
Despite the advances of data collecting and imaging hardware, depth information becomes a major factor restricting the perception of surrounding environment.

In the beginning, researchers use stereo matching algorithms to predict the depth of surrounding environment~\cite{H2007Stereo,yang2018reducing}. Those algorithms are efficient and have acceptable depth estimation results in common scenes, despite the large noise and unstable accuracy~\cite{li2019unconstrained}. Nowadays, convolutional neural networks have been introduced into depth estimation to improve the accuracy of predictions~\cite{eigen2014depth,martins2018fusion}.
Although convolutional neural network introduces a large number of parameters compared with the original matching algorithm, the prediction process can still be performed at a high frame rate thanks to the parallel features of GPUs. Meanwhile, NVIDIA has launched many low-power GPU computing devices, such as NVIDIA Jetson Xavier, NVIDIA Jetson Nano, etc., to further reduce the power consumption of GPU computing. These factors make deep learning-based methods more appealing to be applied to the real world.

On the other hand, the number of public datasets is also rapidly increasing, such as KITTI~\cite{Geiger2013IJRR}, Cityscapes~\cite{2016The}, and nuScenes~\cite{2020nuScenes}.
The variety of public datasets is conducive for training deep neural networks and yielding models that generalize well in real scenes. 
Because researchers generally use rectified cameras to collect datasets, these images usually have a limited field of view.
Under this circumstance, panoramic images are introduced to overcome this limitations for various applications like scene segmentation and depth estimation~\cite{yang2019pass,Jin_2020_CVPR,jiang2021unifuse,yang2021capturing}. The $360^\circ$ surrounding information are projected onto a panoramic image and can be processed at one time, which is efficient for the following steps and upstream applications.

However, most of current algorithms are designed for rectified depth estimation. When the network is transplanted to panoramic images, since lots of features have different degrees of distortion, it poses a great challenge to extract useful features. In addition, since the size of panoramic images is usually much larger, the requirements for GPU resources and speed are also higher. Therefore, an efficient depth estimation network is desired, which should minimize its resource usages while maintaining a high accuracy.

The general learning methods for estimating binocular depth require multiple images for feature matching, which decrease the inference speed in panoramic scenes. Besides, the complexity of current deep neural networks is also increasing, such as ResNet~\cite{he2016deep} and DenseNet~\cite{huang2017densely}. Therefore, using a single panoramic image to estimate the depth, that is, panoramic monocular depth estimation, has its speed advantage. In addition, we consider that the results of basic matching algorithms (such as SGM~\cite{H2007Stereo}, etc.) can be also integrated into the results of panoramic monocular depth estimation to further improve the quality of absolute depth. At the same time, traditional methods and deep learning methods use different hardware to calculate, so there is no conflict during inference.

We extend PADENet~\cite{2020PADENet} first appeared in our previous conference work for outdoor scene understanding, to perform panoramic monocular depth estimation with a focus for indoor scenes. PADENet is adaptable for different scales of distortion in panoramic images. Meanwhile, the network learns both global and local features via corresponding modules. In addition, we improve the training process of the neural network adapted to the characteristics of $360^\circ$ panoramic images. Moreover, we fuse traditional stereo matching algorithm and deep learning methods and further improve the accuracy of depth predictions. An extensive set of experiments demonstrates the effectiveness of our schemes aiming for indoor scene perception. 

The paper's structure is organized as follows: Section 2 gives a review of research works that are useful and related to the panoramic monocular depth estimation. Section 3 details the network structure and methods used in training a high-accuracy and robust depth prediction network. Besides, the fusion strategy for combining traditional and modern approaches are also illustrated. Section 4 presents the results and analysis of the experiments. The conclusion is drawn and the directions of future work are envisioned in the final section.

\section{Related Work}
\subsection{Depth Estimation based on Stereo Matching}
Traditional depth estimation usually uses stereo matching~\cite{H2007Stereo,yang2018reducing,li2019unconstrained}.
After image correction, the two images have matching pixel pairs in the direction of horizontal epipolar, and the deviation between matching pixels represents disparity. According to the binocular camera model, the depth can be calculated through parallax inversion using disparity. There are two common mechanisms for stereo matching, namely sliding window mechanism and energy optimization mechanism. In sliding window mechanism, several methods such as the sum of absolute error (SAD) and the sum of square error (SSD) can be used to measure the similarity of pixel pairs. This matching mechanism is relatively simple, but it lacks accuracy and robustness. The current mainstream method is energy optimization mechanism. The energy function is formed by pixel-matching cost and smoothness cost, which takes into account about global and local features.

A common algorithm for stereo matching is Semi-Global Matching (SGM)~\cite{H2007Stereo}. The algorithm's input is a pair of epipolar-corrected images. The algorithm has four steps: Matching Cost Computation, Cost Aggregation, Disparity Computation, and Disparity Refinement.
Besides, the SGM algorithm~\cite{H2007Stereo} introduces entropy to describe the matching degree of pixel pairs in the images, which improves the robustness in environments such as light changes and texture sparseness.
However, the performance of traditional depth estimation in complex environments still have large room for improvement.

\subsection{Monocular Depth Estimation on Rectified Images}
With the development of convolutional neural network, researches on monocular depth estimation are advancing rapidly.
Eigen et al.~\cite{eigen2014depth} firstly used convolutional neural network for monocular depth estimation by linking Coarse Network and Fine Network to perform the predicting depth.
Fu et al.~\cite{fu2018deep} discretized the depth values into several bins and then performed bin classification instead of direct regression.
Jiao et al.~\cite{jiao2018look} combined monocular depth estimation with semantic segmentation by using an attention-based loss.

Supervised learning can make the convolutional neural networks reach good performance on testing split within the domain seen in the learning process.
But it will also reduce the generalization and robustness in unseen domains due to the bias of datasets. Therefore, unsupervised learning was proposed to make up those deficiencies.
Garg et al.~\cite{zhan2018unsupervised} first proposed an unsupervised training scheme to predict disparity maps, which required left and right views. The disparity maps can be converted to depth maps.
Godard et al.~~\cite{godard2017unsupervised} additionally proposed three types of loss during training: reconstruction loss, image smoothness loss, and left-right consistency loss.
Meanwhile, Luo et al.~\cite{luo2018single} employed deep neural networks to predict the other view image firstly and performed stereo matching afterwards.

\begin{figure*}[!t]
\centering
\fbox{\includegraphics[width=\linewidth]{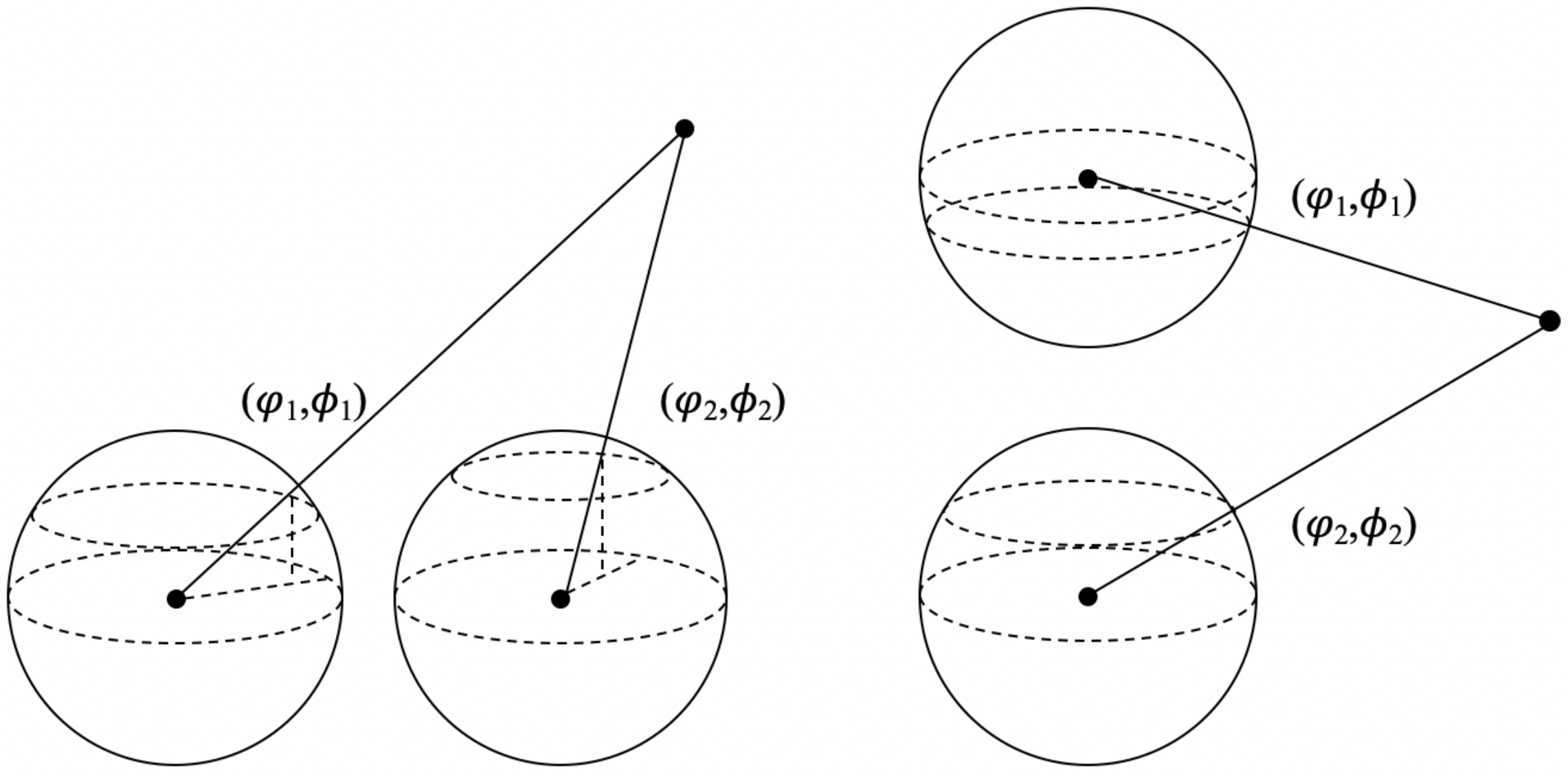}}
\caption{Illustration of two ways of placement of panoramic cameras. On the left is the horizontal placement and on the right is the vertical placement. }
\label{fig:placement}
\end{figure*}

\subsection{Monocular Depth Estimation on Panoramic Images}
Monocular depth estimation has been researched widely in rectified scenes. But when it comes to panoramic scenes, works about monocular depth estimation are much fewer. Gregoire et al.~\cite{payen2018eliminating} proposed a training method with ring padding using the projected KITTI~\cite{Geiger2013IJRR} dataset due to the lack of available panoramic datasets.
Keisuke et al.~\cite{tateno2018distortion} imported distortion-aware convolutions to learn the distorted features in the panoramic images better.
Nikolaos et al.~\cite{zioulis2018omnidepth} put forward UResNet and RectNet by adjusting convolution operation’s receptive field and kernel size in different regions of panoramic images.
Cheng et al.~\cite{2020ODE} designed an omnidirectional depth extension network which embedded a spherical feature transform layer and a deformable convolutional spatial propagation network at the end of feature extraction layers.
Sun et al.~\cite{Sun_2021_CVPR} presented a versatile and efficient framework for holistic understanding of an indoor 360-degree panorama using a Latent Horizontal Feature.
Jin et al.~\cite{Jin_2020_CVPR} proposed a novel learning-based depth estimation framework that leverages the geometric structure of a scene to conduct depth estimation.
Wang et al.~\cite{Wang_2020_CVPR} adopted a two-branch neural network leveraging equirectangular and cubemap projections.
In this research, differing from existing works, we design a new network architecture called PADENet, which is adapted to panoramic features. We leverage different learning methods with PADENet and fuse the outputs with traditional stereo approaches. The experiments demonstrate the effectiveness of our proposed methods.

\section{Approach}
\subsection{Datasets Overview}
Since there are not many real panoramic images collected by cameras at present, many datasets use virtual data.
Currently, three classical indoor datasets are Matterport3D~\cite{2017Matterport3D}, Stanford2D3D~\cite{2017Joint}, and SunCG~\cite{2017Semantic}.
Matterport3D~\cite{2017Matterport3D} contains $10,800$ panoramic images and corresponding depth information; Stanford2D3D~\cite{2017Joint} contains $1,413$ panoramic images, accompanied by depth and normal vectors; SunCG~\cite{2017Semantic} contains $25,000$ panoramic images, but its scene style is different from other datasets.
For investigating the generalization capacity, SunCG~\cite{2017Semantic} is suitable for model evaluation.

3D60~\cite{zioulis2019spherical} is a dataset that contains multi-modal information from real and synthetic large datasets such as Matterport3D~\cite{2017Matterport3D}, Stanford2D3D~\cite{2017Joint}, and SunCG~\cite{2017Semantic}.
It is used for training, testing, and evaluation in this work.
By fusing images from several datasets as in 3D60, the size of the dataset is greatly increased.
At the same time, different datasets represent different types of indoor scenes, so the robustness of the depth estimation model can be also improved during training.
3D60~\cite{zioulis2019spherical} also provides a standard split for training and validation sets, which make it possible to perform comparison and analysis of the results in this work with other existing works.

\subsection{The Settings of Panoramic Cameras}
Deep learning method using convolutional neural network can obtain high-quality prediction results compared to the traditional stereo matching method, but it requires a lot of computing resources.
Therefore, we choose to estimate depth based on a single panoramic image, which saves resources compared to the binocular depth estimation method.
As demonstrated in the field, monocular depth estimation can reach comparative results close to binocular depth estimation with the assistance of convolutional neural networks~\cite{fu2018deep}.

There are two ways to place the cameras for stereo vision: horizontal placement and vertical placement. For rectified images, horizontal placement is generally adopted. According to the epipolar constraint, the matched pixels must be positioned on the same horizontal line. The left and right views are input into the SGM~\cite{H2007Stereo} algorithm to obtain disparity and depth is calculated according to focal length and polar distance. The placement is illustrated in Figure~\ref{fig:placement}.

\begin{figure}[h]
\centering
\fbox{\includegraphics[width=\linewidth]{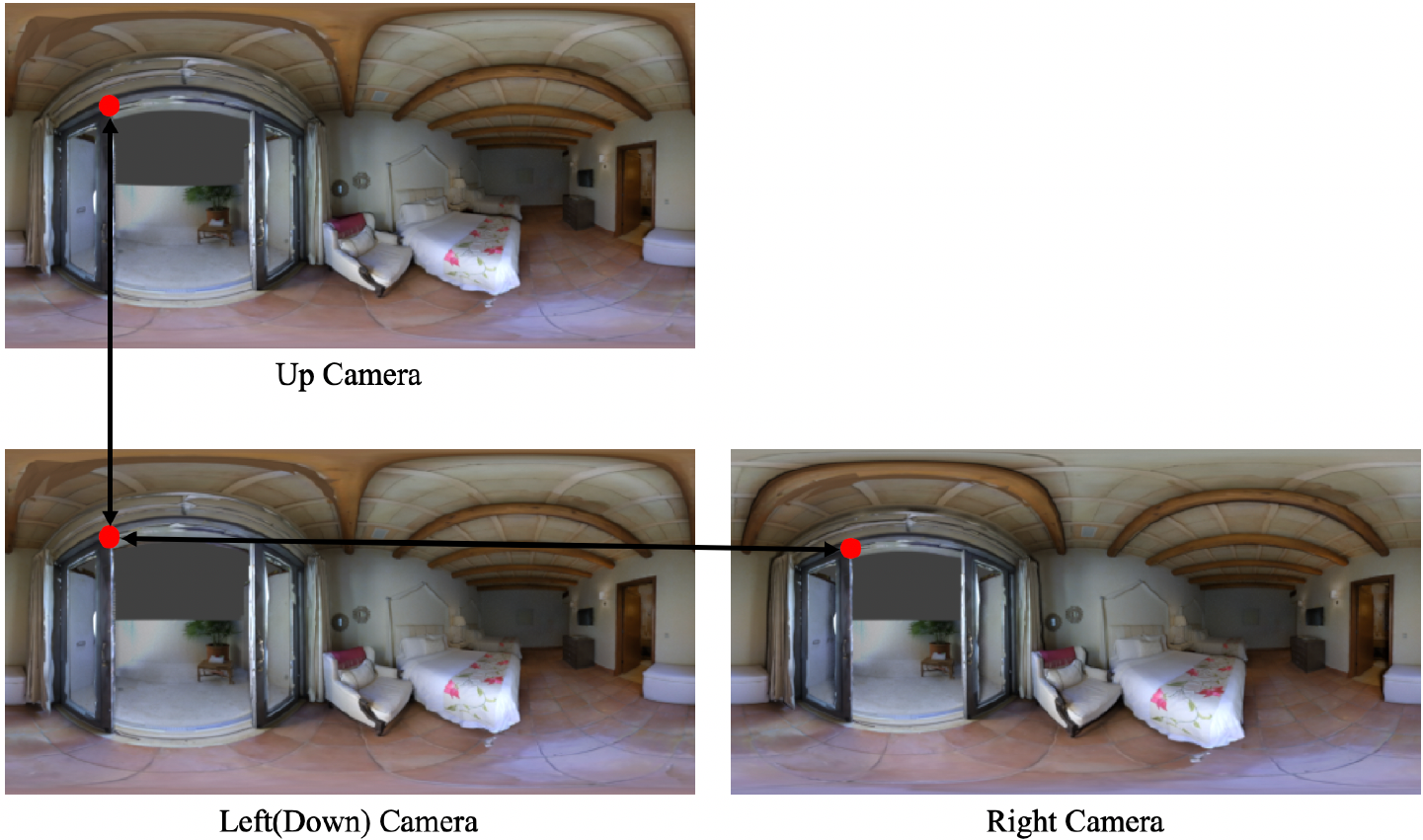}}
\caption{Samples collected by panoramic cameras in different placements. It can be seen that for vertical placement, corresponding points is on the same longitude.}
\label{fig:camera-samples}
\end{figure}

Horizontal placement brings a complex problem when it comes to panoramic images.
This is because the horizontal and vertical coordinates of panoramic images represent the latitude and longitude under spherical coordinate system.
In other words, the epipolar constraint of panoramic images is usually not a straight line in spherical coordinate space. On the contrary, vertical placement is much more beneficial for performing matching across panoramic images, since the epipolar line must be at the same longitude and there is no displacement on the horizontal direction. The latitude difference of the same object in different panoramic cameras indicates the disparity. Then, the depth can be obtained through the depth-disparity relationship in panoramic images as illustrated in Figure~\ref{fig:camera-samples}.

Let's suppose $\textit{x}$ and $\textit{y}$ represent the coordinates in panoramic images.
Besides, $\textit{FOV}_w$ and $\textit{FOV}_h$ represent the view angles of collecting cameras, whereas $\textit{w}$ and $\textit{h}$ represent the shape of panoramic images. Then longitude, denoted as $\phi$, can be written as depicted in Equation~\ref{eq:latitude}.

\begin{equation}
    \phi = \frac{x \cdot FOV_{w}}{w} - \frac{FOV_{w}}{2}
    \label{eq:latitude}
\end{equation}

Similarly, latitude, represented as $\varphi$, can be written as shown in Equation~\ref{eq:longitude}.

\begin{equation}
    \varphi = \frac{y \cdot FOV_{h}}{h} - \frac{FOV_{h}}{2}
    \label{eq:longitude}
\end{equation}

Therefore, the disparity in panoramic images can be written as depicted in Equation~\ref{eq:disparity}.

\begin{equation}
    disparity = \left|\varphi_{1} - \varphi_{2} \right|
    \label{eq:disparity}
\end{equation}

Finally, the corresponding depth is calculated as shown in Equation~\ref{eq:depth} based on the prerequisite baseline length and corresponding latitude.

\begin{equation}
    depth = \frac{baseline \cdot latitude}{disparity}
    \label{eq:depth}
\end{equation}

The $\textit{M}_{angle}$ is shown in Equation~\ref{eq:mangle}, formed by the latitudes of all pixels in the panoramic images.

\begin{equation}
    M_{angle} = \left[
    \begin{array}{ccc}
        cos(FOV_{h} / 2)  & ... & cos(FOV_{h} / 2) \\
        ...               & ... & ... \\
        1                 & ... & 1 \\
        ...               & ... & ... \\
        cos(-FOV_{h} / 2) & ... & cos(-FOV_{h} / 2)
    \end{array}
    \right]
    \label{eq:mangle}
\end{equation}

\begin{figure}[t]
\centering
\fbox{\includegraphics[width=\linewidth]{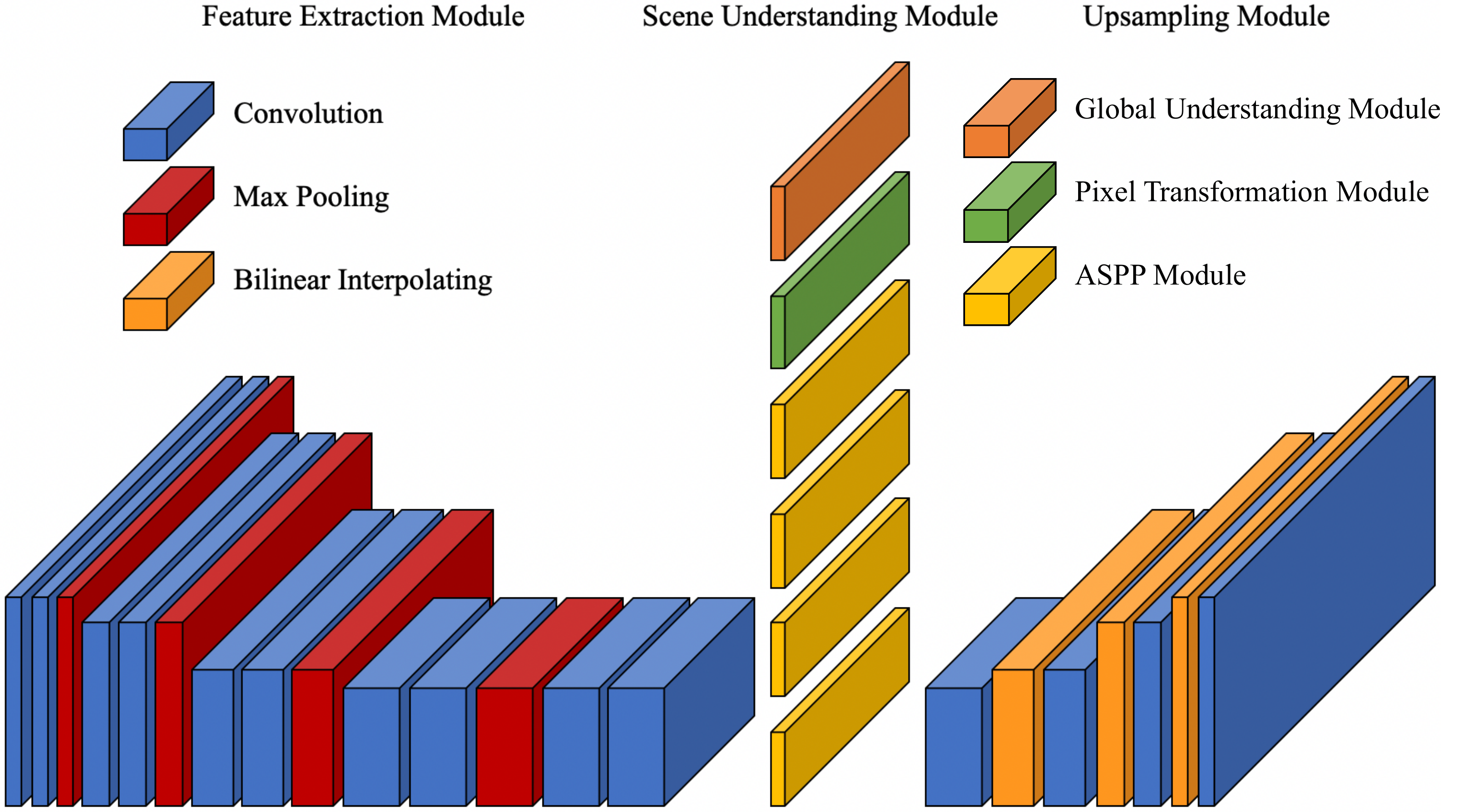}}
\caption{Overview of PADENet architecture. The whole network can be divided into three parts, which are Feature Extraction Module, Scene Understanding Module and Upsampling Module. Convolution operations are represented by blue blocks. Max-pooling operations are represented by red blocks. And bilinear interpolating are represented by yellow blocks.}
\label{fig:padenet}
\end{figure}

\subsection{Monocular Depth Estimation Network Architecture}
Apart from choosing appropriate datasets to train the model, it is also important to design a strong and efficient network structure.
Thereby, we extend the idea of PADENet~\cite{2020PADENet} in our previous conference work and find it still useful in indoor scenes. The network takes a batch of RGB images as inputs and predict corresponding disparity maps as outputs.
The whole structure of the network is presented in Figure~\ref{fig:padenet}.
To make the full use of the information of panoramic images, we design three submodules to extract the features, which are Feature Extraction Module, Scene Understanding Module, and Upsampling Module.
We adopt the VGG~\cite{simonyan2014very} structure to extract initial RGB maps into several tensors.
The following Scene Understanding Module uses a parallel mechanism for learning multi-scale features from the tensors extracted from VGG~\cite{simonyan2014very}.
The detailed model architecture will be shown in the following part. Afterwards, we use bilinear sampling and convolutional layers instead of de-convolutional layers to restore the predictions via the Upsampling Module. Skip-connection mechanism is also added in the network to bridge detail-rich shallow layers and context-critical deep layers for fine-grain prediction.
In the following experiments, we find those designs are beneficial for accurate depth estimation and avoiding excessive calculation.

\begin{figure}[t]
\centering
\fbox{\includegraphics[width=\linewidth]{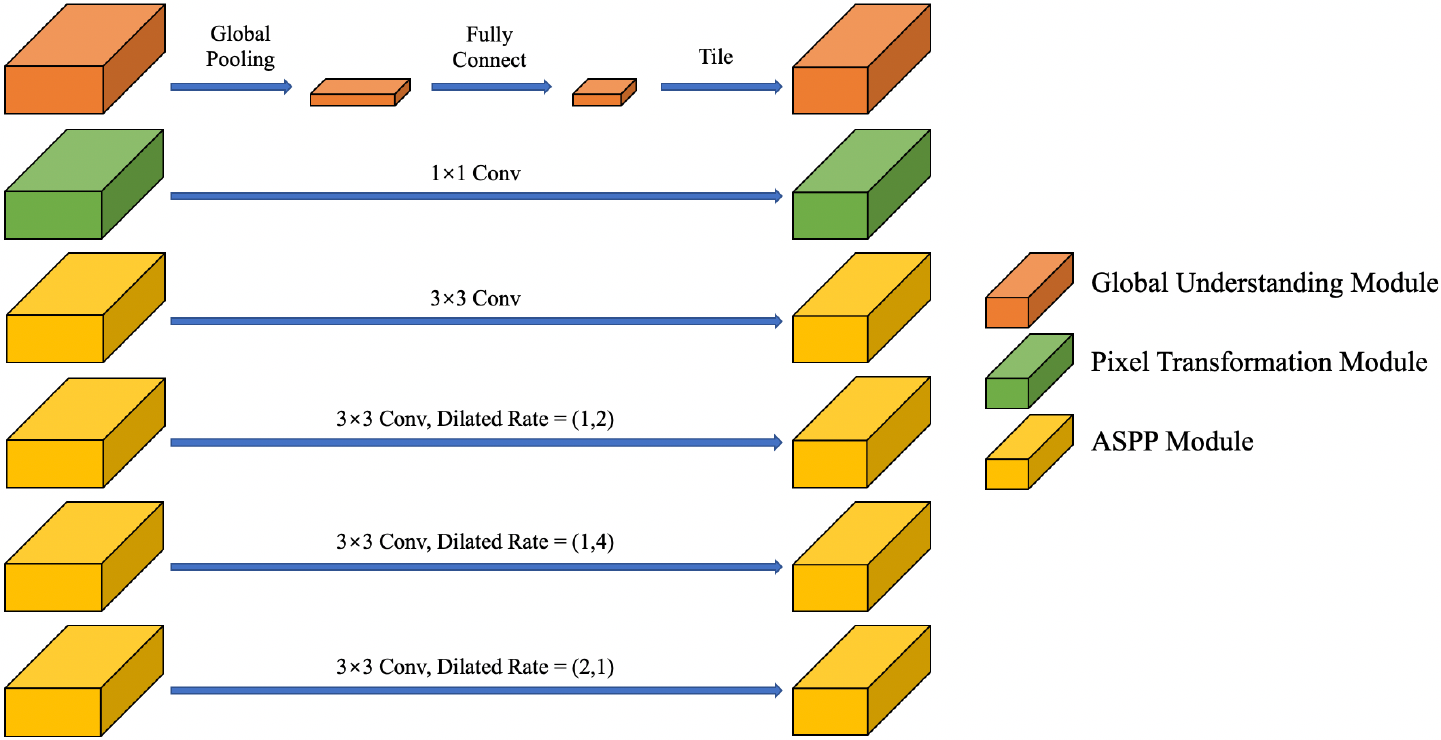}}
\caption{The Scene Understanding Module in PADENet. Global Understanding Module is represented as orange blocks, which uses a global-pooling layer, a fully-connected layer, and a tile operation. Pixel Transformation Module is represented as green blocks, which uses a $1\times1$ convolution. ASPP Module is represented as yellow blocks, which uses dilated convolutions of different dilation rates.}
\label{fig:scene-understanding}
\end{figure}

The detailed design of Scene Understanding Module is inspired by PSPNet~\cite{zhao2017pyramid}, which is shown in Figure~\ref{fig:scene-understanding}.
To adapt the characteristics of panoramic images, the convolutional layers are modified to learn both global and local features and are delivered to the Upsampling Module.
Scene Understanding Module is divided into three parts: Global Understanding Module, Pixel Transformation Module, and Atrous Spatial Pyramid Pooling (ASPP) Module.
In Global Understanding Module, we use global pooling to compress the tensors and add fully-connected layer afterwards.
The results from Global Understanding Module represent the global features of the images.
In Pixel Transformation Module, a $1\times1$ convolutional layer is used to convert each pixel into new values, which addresses pixel-wise learning for specific details.
The last module, the ASPP module, is inspired by the DORN~\cite{fu2018deep} structure. The module contains dilated convolutions with different dilation rates~\cite{chen2017rethinking} to cover different shapes of perceptive regions, which are suitable for different scales of distortion.
Because pole's distortion is much greater than equator's distortion, we set three dilation rates in the horizontal direction and two dilation rates in the vertical direction. The three parts above learn diverse features about panoramic images and preserve enough information.

\subsection{Loss Function}
To train the network for monocular depth estimation, there are mainly two methods: supervised learning and unsupervised learning.
Supervised learning is to constrain the network's outputs to the groundtruth maps, and is suitable for those scenes where real depth data is available in the training split.
However, real depth is usually difficult to obtain for panoramic scenes.
Thereby, unsupervised learning acts as an alternative for panoramic monocular depth estimation.
Unsupervised learning uses binocular images for predicting the disparity maps first, and use an epipolar model to convert disparity to depth later.
We compare the quality of predicting depth using supervised and unsupervised methods and find that the fusion of those two approaches can reach the best performance.

\subsubsection{Supervised Learning}
Supervised learning directly compares the groundtruth disparity map and the predicted disparity map. The difference will then be back-propagated through the network for optimizing the parameters. The most common loss function to describe the difference is L1 loss (absolute difference loss) and L2 loss (squared difference loss). The L1 loss remains a constant slope regardless of the change of loss. The L2 loss decreases the slope when the loss is reduced.

Specifically, L1 loss and L2 loss are depicted in Equation~\ref{eq:l1} and Equation~\ref{eq:l2}.

\begin{equation}
    loss_{l1} = \left|pred-gt\right|
    \label{eq:l1}
\end{equation}

\begin{equation}
    loss_{l2} = \left(pred-gt\right)^2
    \label{eq:l2}
\end{equation}

To combine the advantages of those two loss functions, in this work we choose smooth L1 loss function~\cite{2019IoU}. Smooth L1 loss~\cite{2019IoU} is a piecewise function, which is also inspired by Huber Loss~\cite{laina2016deeper}. As shown in Equation~\ref{eq:smoothl1}, the loss function changes from L1 loss to L2 loss when loss is decreasing. When the loss is large, a certain convergence rate can be maintained to prevent the network from overfitting; when the loss is small, learning rate is reduced for fine-tuning the model.

\begin{equation}
loss_{smoothL1}=
\begin{cases}
delta^2 & \textit{delta $\leq$ 1}\\
delta & \textit{delta > 1}
\end{cases}
\label{eq:smoothl1}
\end{equation}

Here, \textit{delta} represents the absolute difference between the groundtruth and the prediction, as shown in Equation~\ref{eq:delta}.

\begin{equation}
    delta=\left|pred-gt\right|
    \label{eq:delta}
\end{equation}

We also perform loss calculation on different scales of outputs from $4$ up-sampling layers in the network. The multi-scale loss is more effective compared to simply calculating loss on the final level of outputs. The calculation process of multi-scale loss is shown in Equation~\ref{eq:multiscale}.

\begin{equation}
    l_{supervised} = \sum^{4}_{i=1} \frac{1}{4^{4-i}} \cdot loss_{smoothL1}
    \label{eq:multiscale}
\end{equation}

\subsubsection{Unsupervised Learning}
Different from supervised learning, unsupervised learning uses the difference between the real views and the reconstructed views.
Specifically, the more the similar reconstructed views are to the real views, the more precise the predicting disparity maps are. In this work, we use two terms of loss function to describe the loss, as it is shown in Equation~\ref{eq:unsupervised}.

\begin{equation}
    L_{unsupervised} = \sum^{4}_{i=1} \frac{1}{4^{4-i}} \cdot \left(L_{rect}+\lambda_{smooth} L_{smooth} \right)
    \label{eq:unsupervised}
\end{equation}

Reconstruction loss represents the deviation between the real left (right) views and the generated left (right) views, as depicted in Equation~\ref{eq:rect}. The generated views are guided by the predicted disparity maps based on the length of baseline, which is represented by \textit{d}. \textit{F} represents equirectangular projection in Equation~\ref{eq:rect}, which performs transformation from rectangular images to equirectangular images.

\begin{equation}
    L_{rect} = \frac{1}{N}\sum \left|I_{ij} - I_{F \left(F^{-1}  \cdot p_{ij} + \left[d,0,0 \right]^T \right)} \right|
    \label{eq:rect}
\end{equation}

Smoothness loss represents the gradients and smoothness of the predictions, which is shown in Equation~\ref{eq:smooth}. Here, the partial differential values of disparities are multiplied by the corresponding RGB pixel values.

\begin{equation}
    L_{smooth} = \frac{1}{N}\sum \left|\partial_x d_{ij} \right|e^{-\left \|\partial_x I_{ij} \right \|}+\left|\partial_y d_{ij} \right|e^{-\left \|\partial_y I_{ij} \right \|}
    \label{eq:smooth}
\end{equation}

Since we place the cameras vertically, the disparity only has latitude component. When calculating the loss, the epipolar line is still straight in panoramic images. This prerequisite makes the following steps much simpler.

\subsubsection{Fused Learning}
Supervised learning can usually achieve better performance, because it has groundtruth to constrain the training process.
However, supervised learning sometimes will falsely fit the useless feature in the RGB images, which may cause overfitting and lower performance.
On the contrary, unsupervised learning will not be trapped into this problem because it focuses on learning the perspective and occlusion relationship, which is essential for monocular depth estimation.
For example, some features such as decorations on the wall might cause training difficulties in supervised learning, but they can be naturally addressed in unsupervised depth estimation.

In this work, we use the following training strategy.
First, we use unsupervised learning to guide the backward propagation and optimization until the loss is becoming steady.
After this process, the network has a good ability of generalization and global prediction.
Then we perform supervised learning based on the model from unsupervised learning, making the network sensitive to detailed and small objects, and thereby it can achieve a precise depth estimation.

\subsection{Fusion With SGM Results}
A major difficulty for panoramic monocular depth estimation is to predict the absolute depth precisely.
Based on the perspective and occlusion relationship, the network can predict relative depth precisely. But when the network needs to predict depth for new scenes using different cameras, the predicting absolute depth sometimes will have a large deviation.

One feasible solution is to adapt convolutional neural network to binocular depth estimation. Although the network can learn well after training, the requirement of computing resources is much higher than monocular depth estimation since 3D convolution~\cite{2018StereoNet} is needed in the network. Besides, feeding another large-size panoramic image into the network causes the network being heavy and slow.
Another solution is to set an anchor point for the real disparity value. The prediction results of the panoramic monocular depth estimation are then scaled based on the value of the anchor point. The algorithm can make the predictions closer to real situations using additional clues.

\begin{figure}[h]
\centering
\fbox{\includegraphics[width=\linewidth]{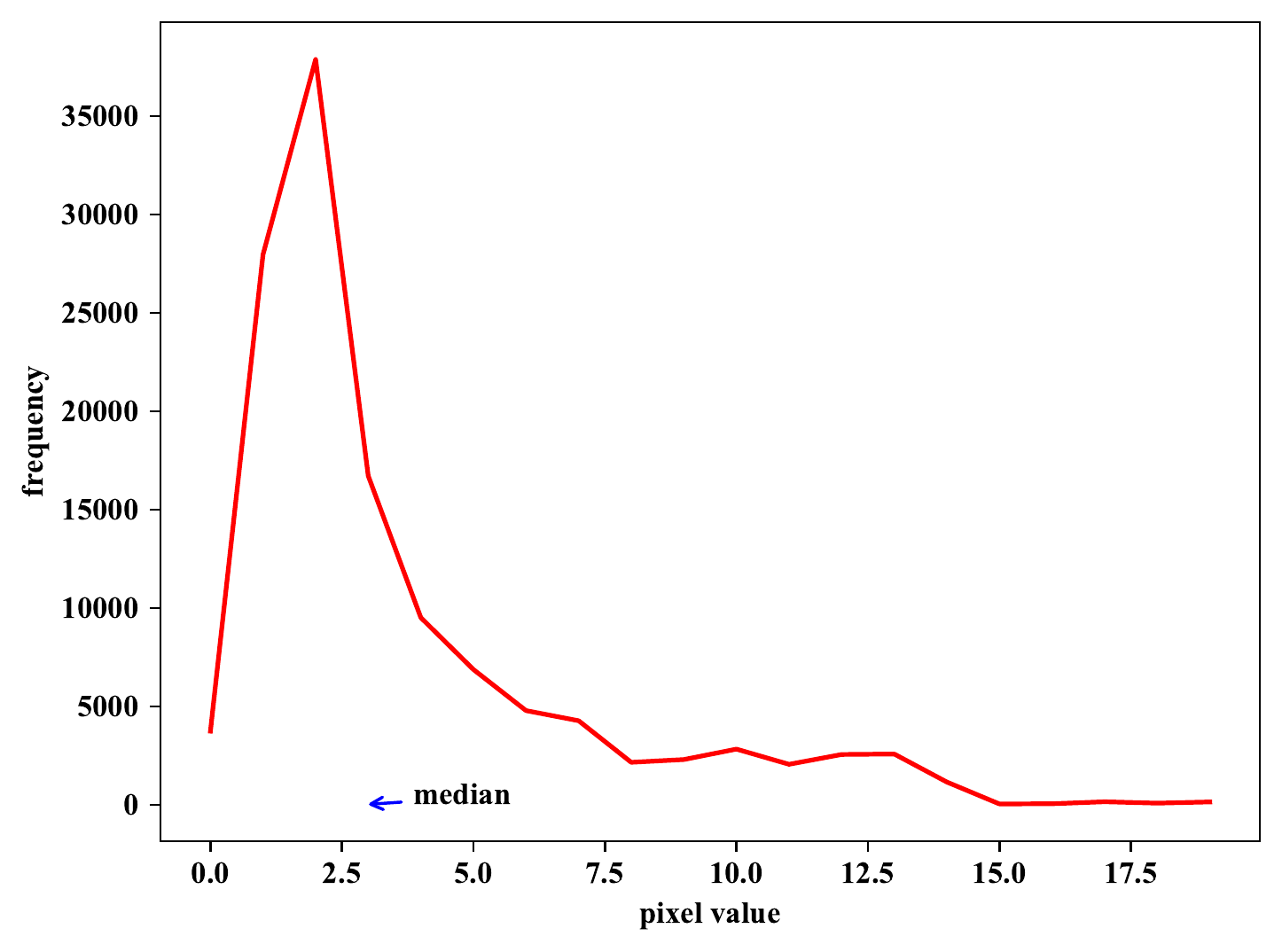}}
\caption{Distribution of depth values predicted by SGM~\cite{H2007Stereo}. The position pointed by the arrow represents the median of all depth values. }
\label{fig:histogram}
\end{figure}

As for the calculation of anchors, we use the outputs of the stereo matching method SGM~\cite{H2007Stereo} for guidance. We choose the median of the SGM~\cite{H2007Stereo} outputs as the anchor as shown in Equation~\ref{eq:sgmscale}, which can reduce the effects of noise and invalidation area and represent the valid value of absolute disparity. Here 
From Figure~\ref{fig:histogram}, we find that the median of disparity is in the valid range. $depth_{network}$ represents the depth predictions from the network. $Median_{network}$ represents the median value of the network outputs and $Median_{SGM}$ represents the median value of SGM~\cite{H2007Stereo} outputs. Both median values are calculated from the entire output panoramas. Experiments show that the strategy can increase the quality of absolute depth predictions.

\begin{equation}
    depth_{final} = depth_{network} \cdot \frac{Median_{network}}{Median_{SGM}}
    \label{eq:sgmscale}
\end{equation}

\section{Experiments and Discussions}
\subsection{Implementation Details}
We use the PyTorch~\cite{paszke2019pytorch} deep learning framework for training.
All training and validation are processed on an NVIDIA GeForce GTX 1080Ti GPU processor. Batch size is set to $2$ and ${\lambda}_{smooth}$ is set to $1$ during training.
We conduct experiments using supervised learning, unsupervised learning, and fused learning.
The network is trained for $20$ epochs and the best performance model is chosen among the checkpoints. Supervised learning is adopted after the loss is steady training with unsupervised learning.
The learning rate is set to $1e-4$ and the Adam optimizer~\cite{kingma2014adam} is utilized during the whole training.

As for dataset, we choose 3D60~\cite{zioulis2019spherical} for training, validation, and testing.
To evaluate the generalization ability, we use the training split of Matterport3D~\cite{2017Matterport3D} in 3D60~\cite{zioulis2019spherical} for training, and the test split of all sub-datasets in 3D60~\cite{zioulis2019spherical} for testing, which covers both real and synthetic scenes, and remains constant in supervised and unsupervised training process.
The training split contains $6826$ images and the testing split contains $325$ images. 

Besides, to evaluate the effectiveness of the fusion with SGM results, we test the model on SunCG~\cite{2017Semantic} because it was collected under different camera settings compared to Matterport3D~\cite{2017Matterport3D}. The whole SunCG~\cite{2017Semantic} contains $754$ images.
All quantitative evaluations are conducted by using the groundtruth available in the 3D60 dataset.
The predicted depth values greater than $20m$ are filtered out.

\begin{figure*}[!t]
\centering
\fbox{\includegraphics[width=\linewidth]{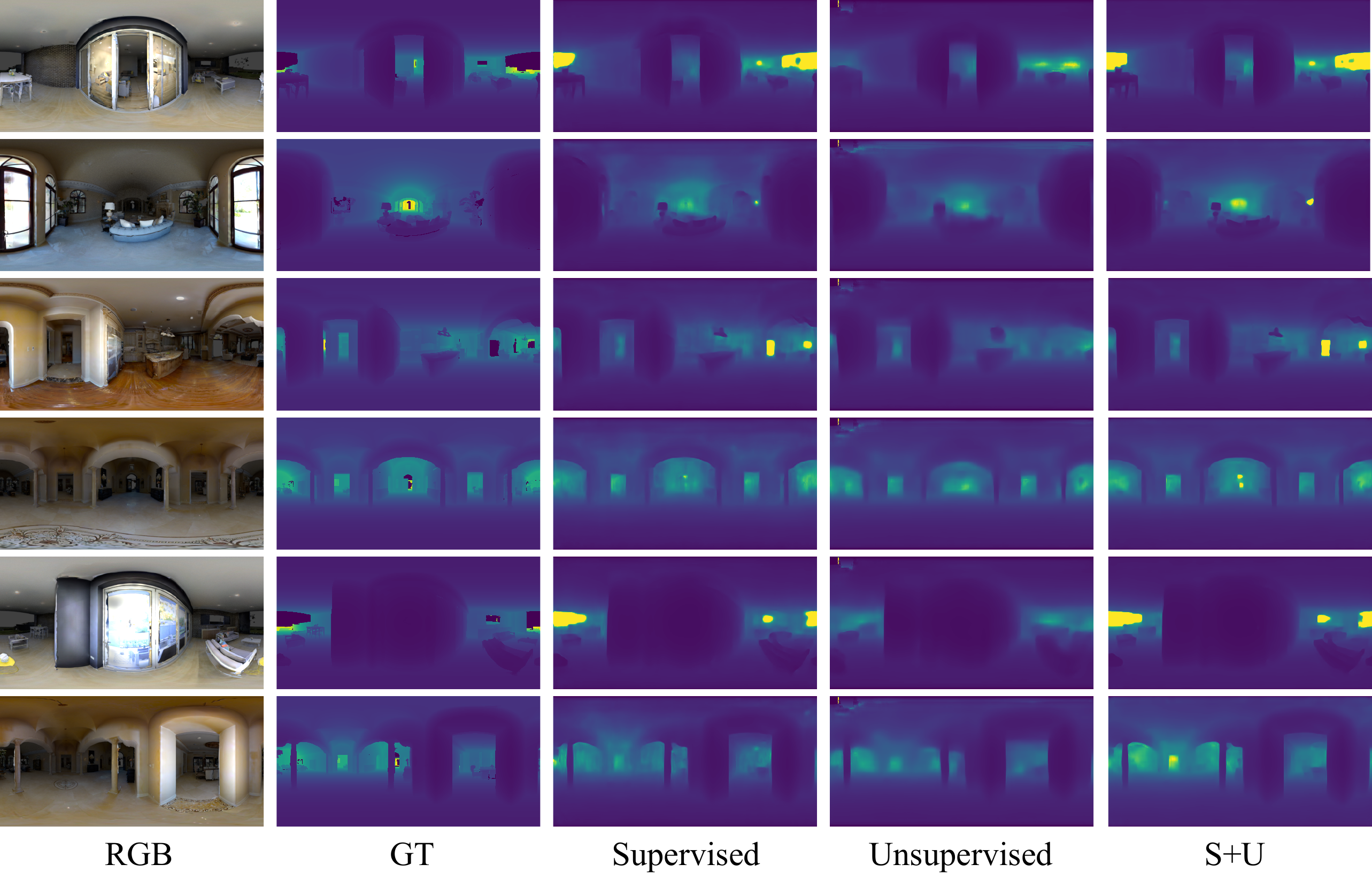}}
\caption{Qualitative examples of different training methods on the test split of 3D60~\cite{zioulis2019spherical}. From left to right are RGB inputs, groundtruths, supervised results, unsupervised results, and fused results. Color is darker nearby and lighter in the distance.}
\label{fig:exp1}
\end{figure*}

\subsection{Model Performance on Different Training Methods}
We conduct experiments on different training methods.
All experiments are based on the PADENet structure.
The first experiment is via supervised learning based on smooth L1 loss.
The second experiment is via unsupervised learning based on reconstruction loss and smoothness loss.
The third experiment is based on the fused learning.

There are mainly four quantitative metrics for evaluating the depth estimation, including Absolute Relative Error (Abs. Rel.), Squared Relative Error (Sq. Rel.), Root Mean Square Error (RMSE and RMSE log), and $\delta$ threshold (Accuracy). These metrics are used when assessing the estimation performance on the testing split.

We compare our model to Nikolaos et al.'s work~\cite{zioulis2019spherical}, which also uses 3D60.
We inherit the mask and sampling strategy which is utilized in~\cite{zioulis2019spherical} for fair comparisons.
The performances of our three models both clearly outperform theirs after evaluating on the testing split.
Despite the model trained via unsupervised learning usually does not behaved as well as the supervised learning one, it still improves the quality of panoramic monocular depth estimation compared to Nikolaos et al.’s work~\cite{zioulis2019spherical}.
And our fused learning achieves the best among the three experimented training methods.
Besides, the inference speed can reach $50$ Frames Per Second (FPS) on an NVIDIA GeForce GTX 1080Ti GPU processor, which can satisfy the requirement of real-time predictions and have the potential to perform real-time panoramic monocular depth estimation in indoor scenes.
Figure~\ref{fig:exp1} shows some examples of depth prediction results of the different training strategies.

\begin{figure*}[!t]
\centering
\fbox{\includegraphics[width=\linewidth]{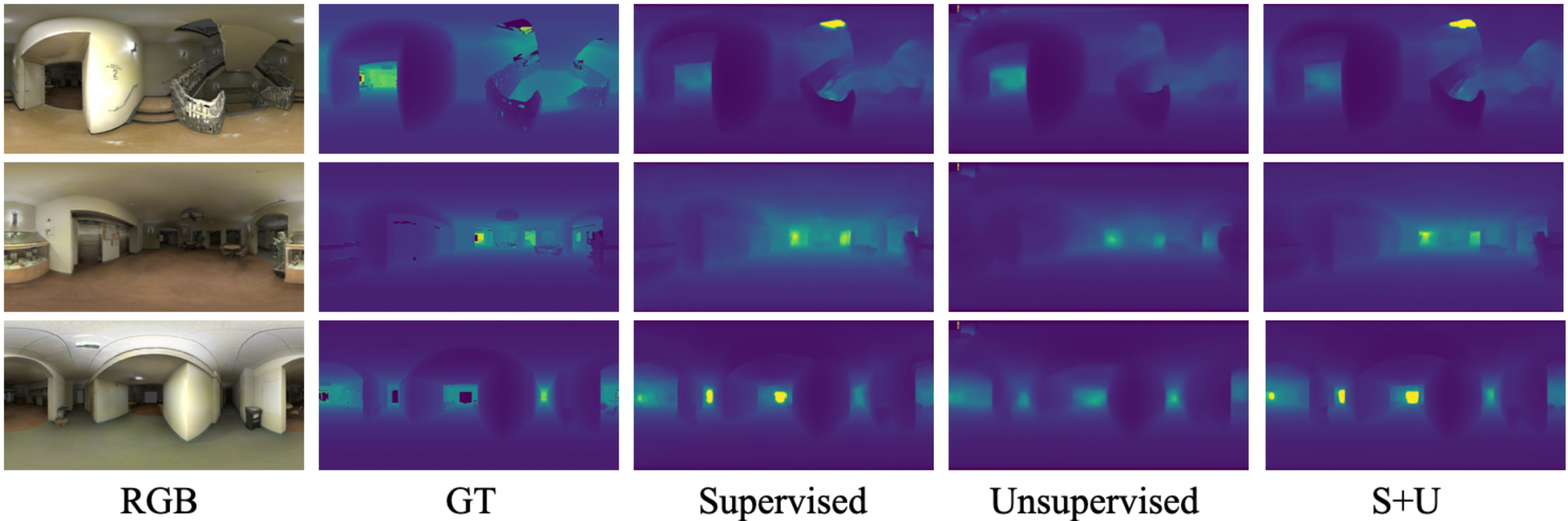}}
\caption{Qualitative examples of different training methods on the test split of Stanford2D3D~\cite{2017Joint}. From left to right are RGB inputs, groundtruths, supervised results, unsupervised results, and fused results. Color is darker nearby and lighter in the distance.}
\label{fig:exp3}
\end{figure*}

To investigate the generalization ability of the model in real-world scenes, we also perform comparisons on Stanford2D3D~\cite{2017Joint}, which are shown in Figure~\ref{fig:exp3}, which verify that the fused learning method achieves robust, fine-grained, and detail-preserved depth estimation on the diverse real indoor panoramas.

\subsection{Quantitative Analysis on Different Training Methods}
There are several quantitative evaluation metrics. Abs. Rel. represents Absolute Relative Error. Sq. Rel. represents Squared Relative Error. RMSE represents Root Mean Squared Error. RMSE log represents Root Mean Squared Error preprocessed by logarithm operation. Acc represents the percentage of the ratio of predicted value to true value within a given range.
 
Through the experimental results, we find that our PADENet is suitable for panoramic monocular depth estimation, either supervised learning or unsupervised learning is adopted.
Experiments show that the results of unsupervised learning are near that of supervised learning.
This indicates that PADENet can learn the effective features of images, and using left and right views as constraints can achieve a similar level compared to supervise learning.
This also explains the possibility that when true depth values are not available for training, unsupervised learning can be used as an alternative for training panoramic monocular depth estimation.

Furthermore, if supervised learning and unsupervised learning are combined, performance on the evaluation metrics can be further improved, surpassing the methods that only use supervised learning or unsupervised learning. First, the model is trained using unsupervised learning to obtain the first-stage model.
The first-stage model can predict the depth distribution of typical objects such as wall, floor and doors.
But for small objects such as sofas and tables, the predictions are relatively rough.
Afterwards, supervised learning is performed to fine-tune those detailed features.
At this point while ensuring the further convergence of global depth estimation, the edge information of objects are much clearer, and the depth information are more precise after fused training.
The detailed numerical comparison is shown in Table~\ref{tab:exp1}.

\begin{table*}[!t]
\centering
\caption{\bf Comparison among Different Training Methods on the Test Split of 3D60~\cite{zioulis2019spherical}.}
\begin{tabular}{cccccccc}
\hline
\textbf{Methods} & \textbf{Abs. Rel.} & \textbf{Sq. Rel.} & \textbf{RMSE} & \textbf{RMSE log} & \textbf{Acc: $\delta < 1.25$} & \textbf{Acc: $\delta < 1.25^2$} & \textbf{Acc: $\delta < 1.25^3$} \\
\hline
Nikolaos et al.~\cite{zioulis2019spherical} & 0.138 & 0.091 & 0.473 & 0.184 & 82.4$\%$ & 95.9$\%$ & 98.9$\%$ \\
Supervised Learning & 0.061 & 0.036 & 0.366 & 0.132 & 91.3$\%$ & 98.5$\%$ & 99.5$\%$ \\
Unsupervised Learning & 0.070 & 0.043 & 0.402 & 0.177 & 88.7$\%$ & 97.4$\%$ & 99.0$\%$ \\
Fused Learning & 0.050 & 0.027 & 0.326 & 0.121 & 94.0$\%$ & 98.9$\%$ & 99.7$\%$ \\
\hline
\end{tabular}
\label{tab:exp1}
\end{table*}

\subsection{Effectiveness of Fusing SGM Results}
\begin{figure*}[!t]
\centering
\fbox{\includegraphics[width=\linewidth]{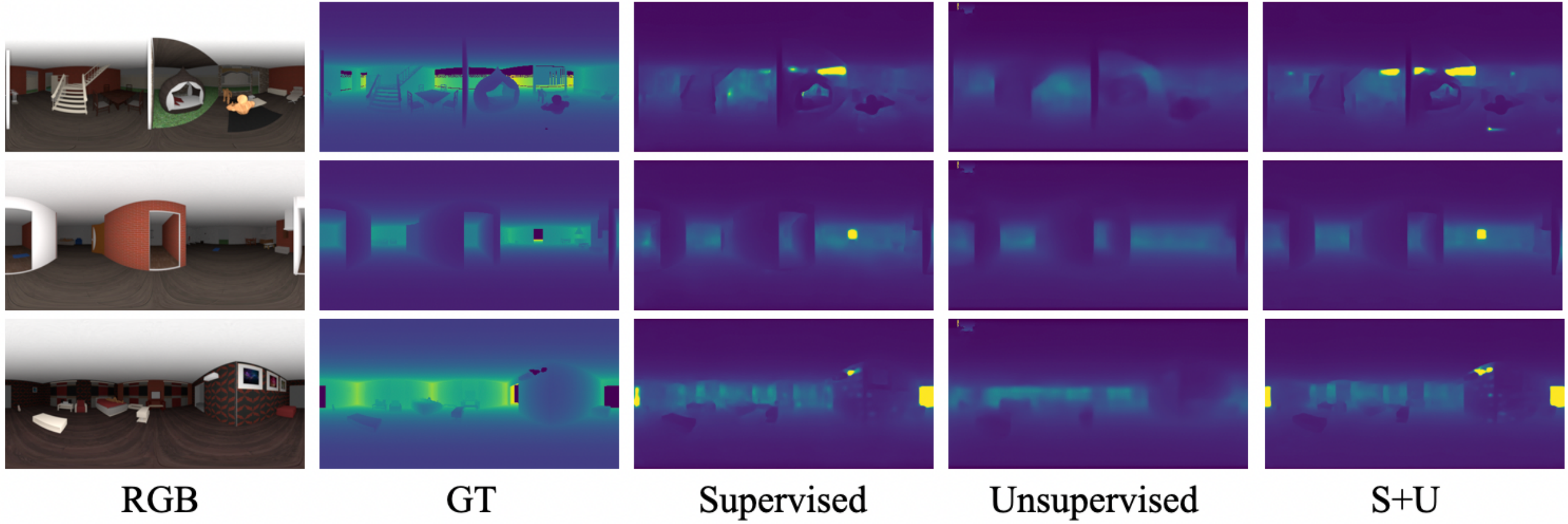}}
\caption{Qualitative examples of fusing SGM results on SunCG~\cite{2017Semantic}. From left to right are RGB inputs, groundtruths, supervised results, unsupervised results, and fused results. Color is darker nearby and lighter in the distance.}
\label{fig:exp2}
\end{figure*}

As mentioned above, the training model on one dataset often does not fit well to another different scene using different camera setups when predicting the absolute depth.
In this work, we also add the anchor points, which are the median of SGM~\cite{H2007Stereo} outputs, to the initial model.
We test our idea on SunCG~\cite{2017Semantic}, a different dataset.
The results in Table~\ref{tab:exp2} show that the SGM-fused strategy can improve the absolute depth predictions, especially on RMSE and Accuracy.
The RMSE and Accuracy metrics describe the absolute deviation and quality of predictions.
From Table~\ref{tab:exp2}, we find that the results on those metrics have more than $20\%$ increase, which proves the effectiveness of the proposed fusion method.
The visualization results in Figure~\ref{fig:exp2} also verify the superiority of the prediction fusion strategy, which helps to attain more reliable depth estimation.

\begin{table*}[!t]
\centering
\caption{\bf Comparison among whether SGM results are utilized on SunCG~\cite{2017Semantic}.}
\begin{tabular}{cccccccc}
\hline
\textbf{Methods} & \textbf{Abs. Rel.} & \textbf{Sq. Rel.} & \textbf{RMSE} & \textbf{RMSE log} & \textbf{Acc: $\delta < 1.25$} & \textbf{Acc: $\delta < 1.25^2$} & \textbf{Acc: $\delta < 1.25^3$} \\
\hline
\multicolumn{8}{c}{Supervised Learning with SGM} \\
\hline
w/o SGM & 0.167 & 0.190 & 0.826 & 0.294 & 38.6$\%$ & 72.0$\%$ & 97.4$\%$ \\
w SGM & 0.167 & 0.239 & 0.800 & 0.252 & 50.3$\%$ & 89.4$\%$ & 96.6$\%$ \\
\hline
\multicolumn{8}{c}{Unsupervised Learning with SGM} \\
\hline
w/o SGM & 0.187 & 0.223 & 0.936 & 0.348 & 31.8$\%$ & 62.2$\%$ & 94.0$\%$ \\
w SGM & 0.183 & 0.240 & 0.841 & 0.298 & 42.8$\%$ & 79.7$\%$ & 95.5$\%$ \\
\hline
\multicolumn{8}{c}{Fused Learning with SGM} \\
\hline
w/o SGM & 0.162 & 0.178 & 0.812 & 0.287 & 39.5$\%$ & 73.7$\%$ & 97.9$\%$ \\
w SGM & 0.163 & 0.223 & 0.771 & 0.246 & 51.3$\%$ & 90.1$\%$ & 96.9$\%$ \\
\hline
\end{tabular}
\label{tab:exp2}
\end{table*}

\section{Conclusion and Future Work}
Panoramic images can greatly increase the horizon for visual sensors to observe and collect data.
But its distortion has limited the application of the collected panoramic data.
Monocular depth estimation, as a new approach to obtain the depth values through a single camera, has drawn much attention recently due to the rapid development of convolutional neural networks and deep learning.
We extend our idea of PADENet~\cite{2020PADENet}, which first appeared in our previous conference work for outdoor road-driving images, to indoors scene understanding.
We further fuse supervised learning and unsupervised learning methods in our experiments and propose a learning regimen to train the network.
Besides, we also combined the traditional SGM~\cite{H2007Stereo} outputs with the network's outputs and improve the network's generalization ability.

There are still much space for improvement in the future.
Although PADENet~\cite{2020PADENet} can infer at a real-time frame rate, more efficient backbones can be applied if larger images are fed into the network. Besides, fusing SGM~\cite{H2007Stereo} outputs to the network can be further explored with more effective and sophisticated strategies to advance the quality and the robustness of predictions.
Although the variety and complexity of training datasets are growing rapidly to make the network adaptable to most scenes, domain adaptation is still a promising technology when scenes contain transparent~\cite{zhang2021trans4trans} or reflective objects~\cite{hong2021panoramic} or are under different intensities of illumination~\cite{vankadari2020unsupervised,albanis2021pano3d}.
We will continuously improve our model to make it more fit and robust to the panoramic scenes and try to close the gap between indoor scenes and outdoor scenes.

\section*{Funding}
This research was granted from ZJU-Sunny Photonics Innovation Center (No. 2020-03). This research was also funded in part through the AccessibleMaps project by the Federal Ministry of Labor and Social Affairs (BMAS) under the Grant No. 01KM151112, and in part by the University of Excellence through the ``KIT Future Fields'' project.

\section*{Acknowledgments}
This research was supported in part by Hangzhou SurImage Technology Company Ltd.

\section*{Disclosures}
The authors declare that there are no conflicts of interest related to this article.
%%The authors declare no conflicts of interest.

% Bibliography
\bibliography{mybib}

% Full bibliography added automatically for Optics Letters submissions; the following line will simply be ignored if submitting to other journals.
% Note that this extra page will not count against page length
% \bibliographyfullrefs{mybib}

%Manual citation list
%\begin{thebibliography}{1}
%\bibitem{Zhang:14}
%Y.~Zhang, S.~Qiao, L.~Sun, Q.~W. Shi, W.~Huang, %L.~Li, and Z.~Yang,
 % \enquote{Photoinduced active terahertz metamaterials with nanostructured
  %vanadium dioxide film deposited by sol-gel method,} Opt. Express \textbf{22},
  %11070--11078 (2014).
%\end{thebibliography}

% Please include bios and photos of all authors for aop articles
\ifthenelse{\equal{\journalref}{aop}}{%
\section*{Author Biographies}
\begingroup
\setlength\intextsep{0pt}
\begin{minipage}[t][6.3cm][t]{1.0\textwidth} % Adjust height [6.3cm] as required for separation of bio photos.
  \begin{wrapfigure}{L}{0.25\textwidth}
    \includegraphics[width=0.25\textwidth]{john_smith.eps}
  \end{wrapfigure}
  \noindent
  {\bfseries John Smith} received his BSc (Mathematics) in 2000 from The University of Maryland. His research interests include lasers and optics.
\end{minipage}
\begin{minipage}{1.0\textwidth}
  \begin{wrapfigure}{L}{0.25\textwidth}
    \includegraphics[width=0.25\textwidth]{alice_smith.eps}
  \end{wrapfigure}
  \noindent
  {\bfseries Alice Smith} also received her BSc (Mathematics) in 2000 from The University of Maryland. Her research interests also include lasers and optics.
\end{minipage}
\endgroup
}{}

\end{document}